\documentclass[conference]{IEEEtran}
\usepackage{cite}
\usepackage{amsmath,amssymb,amsfonts}
\usepackage{algorithm}
\usepackage{algpseudocode}
\usepackage{microtype}
\usepackage{graphicx}
\usepackage{textcomp}
\usepackage{xcolor}
\usepackage{hyperref}
\usepackage{float}
\usepackage{siunitx}
\def\BibTeX{{\rm B\kern-.05em{\sc i\kern-.025em b}\kern-.08em
    T\kern-.1667em\lower.7ex\hbox{E}\kern-.125emX}}

\makeatletter
\newcommand{\linebreakand}{%
  \end{@IEEEauthorhalign}
  \hfill\mbox{}\par
  \mbox{}\hfill\begin{@IEEEauthorhalign}
}
\makeatother

\begin{document}

\title{TinyNav: End-to-End TinyML for Real-Time Autonomous Navigation on Microcontrollers}

\author{

\IEEEauthorblockN{Pooria Roy}
\IEEEauthorblockA{
    \textit{Queen's University} \\
    pooria.roy@queensu.ca
}

\and
\IEEEauthorblockN{Nourhan Jadallah}
\IEEEauthorblockA{
    \textit{Queen's University} \\
    23jpv1@queensu.ca
}

\and
\IEEEauthorblockN{Tomer Lapid}
\IEEEauthorblockA{
    \textit{Queen's University} \\
    23rtn2@queensu.ca
}

\and
\IEEEauthorblockN{Shahzaib Ahmad}
\IEEEauthorblockA{
    \textit{Queen's University} \\
    shahzaib.ahmad@queensu.ca
}

\linebreakand 
\IEEEauthorblockN{Armita Afroushe}
\IEEEauthorblockA{
    \textit{Queen's University} \\
    24kb11@queensu.ca
}

\and
\IEEEauthorblockN{Mete Bayrak}
\IEEEauthorblockA{
    \textit{Queen's University} \\
    mete.bayrak@queensu.ca
}

} 
\maketitle
\begin{abstract}
Autonomous navigation typically relies on power-intensive processors, limiting accessibility in low-cost robotics. Although microcontrollers offer a resource-efficient alternative, they impose strict constraints on model complexity. We present TinyNav, an end-to-end TinyML system for real-time autonomous navigation on an ESP32 microcontroller. A custom-trained, quantized 2D convolutional neural network processes a 20-frame sliding window of depth data to predict steering and throttle commands. By avoiding 3D convolutions and recurrent layers, the 23k-parameter model achieves 30 ms inference latency. Correlation analysis and Grad-CAM validation indicate consistent spatial awareness and obstacle avoidance behavior. TinyNav demonstrates that responsive autonomous control can be deployed directly on highly constrained edge devices, reducing reliance on external compute resources. Code and schematics are available at: \url{https://github.com/regularpooria/tinynav}.
\end{abstract}


\section{Introduction}
A central challenge in current low-cost robotics is the capacity for autonomous, real-time obstacle avoidance and path planning. Existing solutions often rely on manual control, rigid predetermined algorithms, or computationally expensive single-board computers that require involved algorithms such as Simultaneous Localization and Mapping (SLAM). This project aims to bridge this gap by leveraging recent advancements in 32-bit microcontroller hardware and edge machine learning frameworks, specifically TensorFlow Lite (TFLite) for Microcontrollers \cite{david2021tensorflowlitemicroembedded} and ESP-NN \cite{espressif2026espnn}. Specifically, we demonstrate the practicality of deploying quantized Convolutional Neural Network (CNN) models on low-resource microcontrollers to process digital spacial data from multi-zone Time-of-Flight (ToF) sensors. By enabling real-time autonomous navigation on these constrained platforms, this approach significantly lowers the barrier to entry for intelligent robotics, shifting sophisticated, adaptive behaviors away from expensive processors and directly onto cost-effective edge devices.

\subsection{Motivation}

Tiny Machine Learning (TinyML) is a subfield of AI that integrates machine learning models with embedded systems, enabling deployment on low-power devices such as microcontrollers and sensors. It supports real-time data processing while improving privacy, latency, energy efficiency, and reducing dependence on cloud infrastructure, making it well-suited for resource-constrained environments. Between 2020 and 2024, TinyML research grew exponentially, averaging a 59.23\% annual increase in publications across Electrical and Electronics Engineering, Computer Science and Information Systems, and Telecommunications \cite{han2026tinyml}. Future developments target advanced multi-modal sensor integration within constrained hardware and the edge-to-cloud continuum, where computationally intensive tasks such as model training are offloaded to the cloud while real-time inference runs locally to maintain responsiveness.

\begin{figure}[H] 
\centering
\includegraphics[width=\columnwidth]{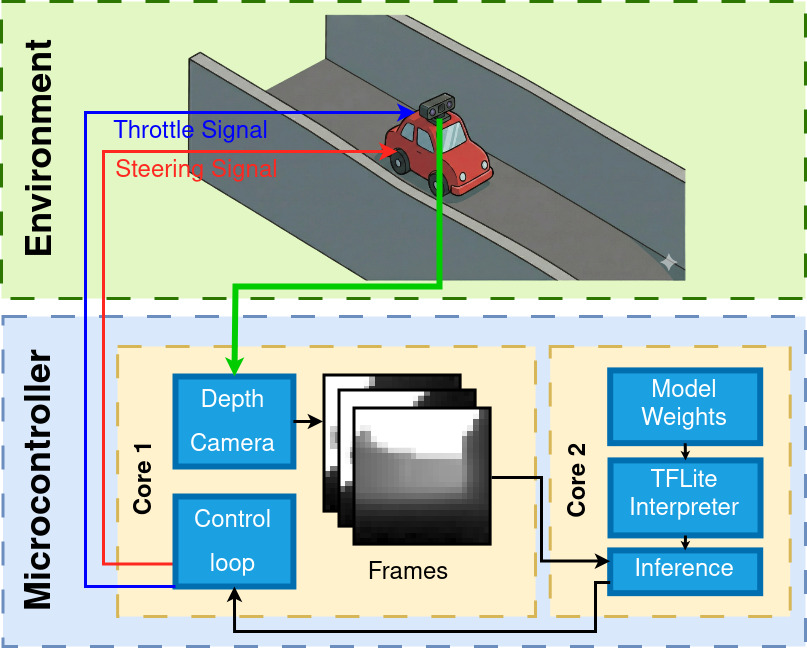}
\caption{TinyNav Real‑Time Autonomous Navigation Pipeline}
\label{fig:tinynav_pipeline_hero}
\end{figure}

TinyML enables fully on-board AI in systems ranging from drones to medical devices. This work investigates deploying a depth camera with a relatively large AI model on an ESP32 microcontroller for real-time autonomous navigation. By extending existing TinyML integration frameworks through parallel-processing shown in Figure \ref{fig:tinynav_pipeline_hero}, the proposed system contributes to intelligent, resource-efficient applications across robotics, healthcare, and industrial automation.

\subsection{Related Works}

The deployment of machine learning models on memory-constrained edge devices necessitates aggressive optimization strategies, primarily focusing on model compression, kernel optimizations, and custom hardware such as Neural Processing Units (NPUs)

\subsubsection{Pruning \& Quantization: }
To deploy CNNs on memory-constrained microcontrollers, pruning and quantization are widely used \cite{ARIF202513}. Pruning techniques, including static and activation-based methods, remove redundant weights or neurons to reduce memory usage, typically with minor accuracy loss. This is often necessary when SRAM limits prevent using the original model. Quantization further compresses models by converting FP32/FP16 weights to lower-precision formats such as INT8 or INT1. It is also required by frameworks like TFLite Micro for integer-optimized hardware, enabling efficient execution with minimal performance degradation \cite{ARIF202513}.

\subsubsection{Real-Time Inference Constraints}
Beyond memory limits, inference speed remains a central bottleneck for autonomous robotics, where delayed decision-making can result in collisions. Prior research mentions that it is possible to run a 7-layer CNN network with a small backbone at 30 frames per second \cite{ARIF202513}. Even though bigger models can fit in the microcontroller's RAM, inference time is a critical aspect in autonomous driving, and any high inference latency will cause significant lag and result in a lazy response.

\subsubsection{TinyML Frameworks and Sensor Dependency}
A recent 2023 survey on TinyML outlines various pre-developed architectures for object detection that can serve as robust backbone networks for edge applications \cite{10177729}. However, adapting these models reveals significant hardware dependencies. For instance, recent foundational work on TinyML for speech recognition successfully deployed a 1D CNN for keyword spotting using the Edge Impulse framework \cite{Barovic_2025}. While this specific implementation bypassed initial pruning and quantization, the authors acknowledged these compression techniques as necessary next steps for scaling complexity. Crucially, this prior research highlighted extreme sensor sensitivity: achieving high accuracy mandated that the training data be captured using the exact same physical sensor hardware as the deployment environment \cite{Barovic_2025}. This hardware-software coupling is highly relevant for spatial navigation systems, strongly suggesting that the ToF sensor configurations used during the training phase must perfectly mirror the live inference environment to prevent severe degradation in real-world accuracy.

\subsection{Problem Definition}

The integration of machine learning capabilities directly within edge devices has emerged as a rapidly growing research area. However, this paradigm is inherently constrained by severe hardware limitations, including restricted computational power, limited memory capacity, and reduced energy availability. Although TinyML architectures can achieve extremely low inference latencies (0-5ms) they remain incapable of supporting highly complex neural networks, such as Large Language Models (LLMs), or accommodating extensive computational libraries and advanced processing functions such as 3D Convolutional Layers (Conv3D), Long Short-Term Memory (LSTM), Gated Recurrent Units (GRU), and Multi-Head Attention (MHA). These constraints significantly limit the deployment of sophisticated AI models on microcontrollers.

To address the inherent limitations of microcontroller-based systems, particularly the constrained memory and processing capacity of the ESP32, this research defines a structured methodology that accounts for these constraints throughout the study. The approach relies on custom data collection to control input size and complexity, model quantization to reduce memory usage and computational load, a compact CNN for embedded deployment, and parallel processing to make effective use of the device’s dual-core capabilities and to isolate control-loop latencies from model performance. Based on these constraints, the study adopts the following process to design a custom robot:

\begin{enumerate}
    \item Robot hardware design and assembly.
    \item Sensor calibration and component validation.
    \item Dataset collection using depth camera frames.
    \item Data pre-processing and sanitization to remove noise and irrelevant inputs.
    \item Training a CNN network to predict steering and throttle values.
    \item Model compression and deployment on the ESP32 achieved strictly through quantization to optimize the model footprint.
    \item Performance optimization and output error minimization to bridge simulation-to-reality discrepancies.
\end{enumerate}

\section{Methodology}

\subsection{Components}
\subsubsection{Microcontroller ESP32-P4-WIFI6-M}
The selection of an appropriate microcontroller is fundamental to achieving real-time inference on resource-constrained platforms. For this to work, we deployed the Waveshare ESP32-P4-WIFI6 development board, a dual chip architecture combining ESP32-P4 primary processor with an ESP32-C6 wireless coprocessor operating at 360MHz with 32 MB integrated PSRAM, 768 KB L2 memory, and 32 MB onboard NOR flash. The platform integrated dedicated hardware accelerators including JPEG codec, Pixel Processing Accelerator (PPA), Image Signal Processor (ISP), and H.264 encoder. Wireless connectivity is provided through an ESP32-C6-MINI module supporting Wi-Fi 6 (2.4 GHz) and Bluetooth 5.3 LE, communicating via SDIO interface to prevent radio operations from interfering with inference timing. At approximately \~\$20 USD, the platform balances cost with computational capability for embedded AI applications.
\subsubsection{ToF Depth Camera Sipeed MaixSense A010}
Spatial perception was achieved using the Sipeed MaixSense A010, a Time-of-Flight (ToF) depth sensor providing structured $100\times100$ pixel depth measurements at up to 20 FPS. The A010 outputs quantized depth values directly via UART at 115200 baud. The sensor operates at 940 nm infrared wavelength with a $\ang{70} \times \ang{60}$ field of view and an effective range of 200-2500 mm.

The A010 uses adaptive quantization to compress 16-bit internal measurements into 8-bit transmission values. We selected linear mode (distance = UNIT+10), providing uniform 10 mm precision across the full sensing range. Reverse engineering of the UART protocol revealed discrepancies with manufacturer documentation. Resolution is dynamically reported in the header, requiring runtime detection rather than compile-time constants. For our use case applying an on sensor $4\times4$ binning to reduce native resolution from $100\times100$ to $25\times25$ pixels helped to keep information while reducing input size and therefore memory size.
\subsubsection{Robot Design Principles}

In designing the robot, we deliberately selected readily available components sourced from Amazon in order to make the project easier to replicate and more cost-effective for future implementations. This approach reduces barriers to entry for other researchers or practitioners who may wish to reproduce or extend the system, while also minimizing overall development time and procurement complexity.

For the driving mechanism, we implemented a tank drive configuration to reduce the number of control variables that the learning model must account for, particularly when compared to conventional car like steering systems that require consideration of turning radius, steering geometry, and clearance constraints during maneuvering. The tank drive system enables the robot to rotate in place, which simplifies motion planning and provides greater positional flexibility in confined environments. Additionally, because this configuration does not rely on a front wheel steering assembly, it allows for a more compact track width and improved space efficiency, which is advantageous in environments with limited clearance.
\subsection{Data}
To train the navigation model, we collected a custom data set using the ToF sensor on the robot. The camera streamed frames at 20 FPS, and each frame was paired with the steering and throttle commands sent to the motors at that moment. This created a labeled dataset that maps depth images directly to control outputs. 
Each sample consists of three components: an image, a steering value, and a throttle value. Since recurrent architectures such as LSTM are not efficiently supported on the ESP32, temporal information was incorporated using a sliding window approach. For each training instance, 20 consecutive frames (approximately one second of visual history at 20 FPS) were stacked along the channel dimension to form a single input tensor. This produces a 20-channel input that allows a 2D convolutional neural network to learn short-term patterns without relying on recurrent layers. Frames were rotated to ensure consistent orientation across all recordings. Images were resized from 25$\times$25 to 24$\times$24 to better align with embedded convolution optimizations that favor dimensions divisible by two. Horizontally flipped samples were generated to improve robustness and reduce directional bias. The dataset was shuffled across different tracks to prevent sequence bias, and an 60/40 train/test split was used to evaluate generalization performance. This pre-processing pipeline produces compact, temporally-aware inputs suitable for real-time inference on a resource-constrained microcontroller.

\begin{figure}[h]
    \centering
    \includegraphics[width=0.5\linewidth]{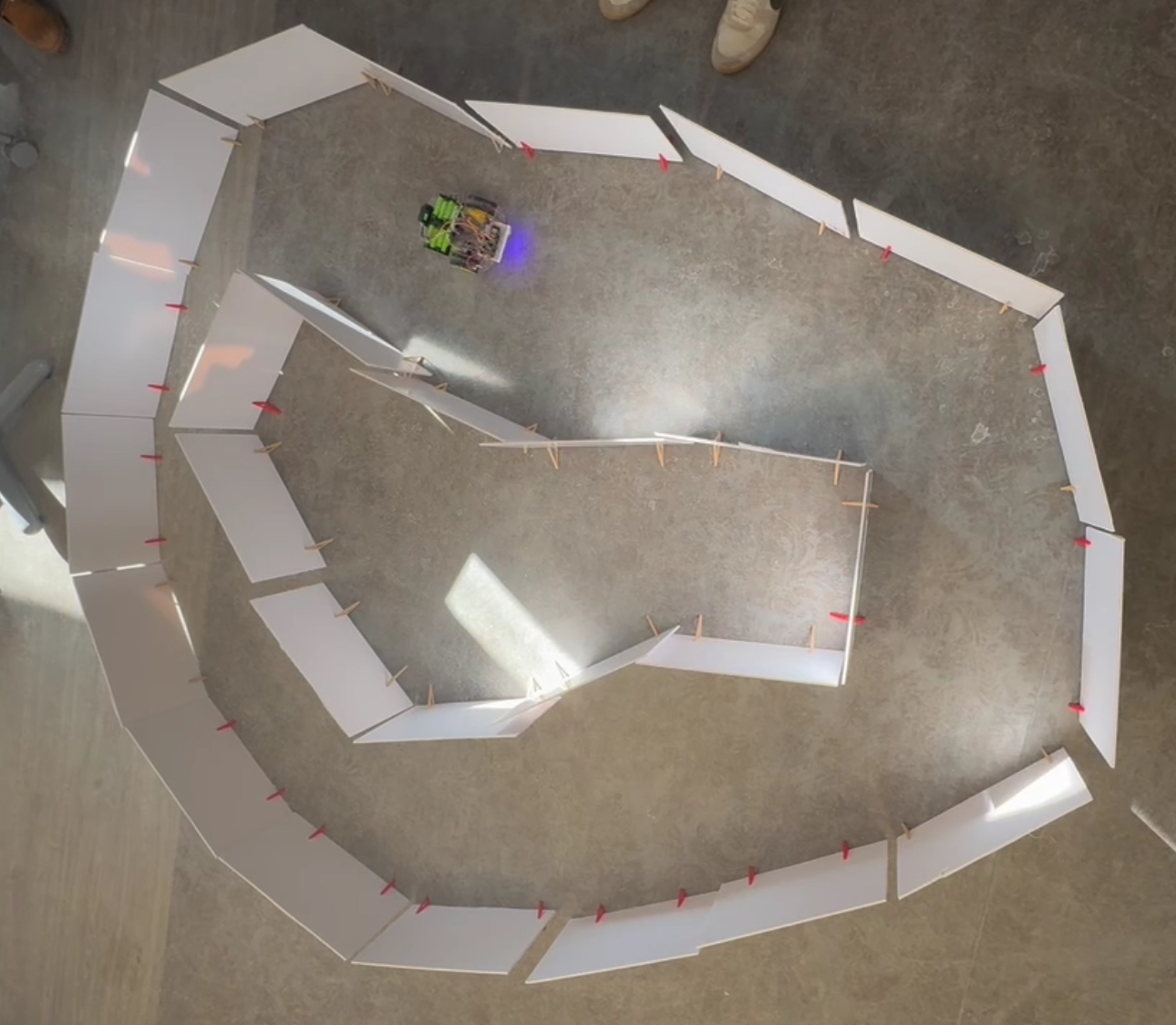}
    \caption{Example: top-down view of a track}
    \label{fig:tinynav_route_example}
\end{figure}

Data were recorded across multiple hand-built track layout constructed from modular wall segments. The track layout was modified between sessions by changing the wall spacing, introducing non-parallel walls, and rearranging corridor segments. Narrow passages and sharp turns were intentionally included to force precise steering corrections. Data were also recorded on different ground surface materials to account for traction variations and motion noise. By varying the track geometry during collection, the model was exposed to a wider range of spatial patterns rather than memorizing a single layout. Figure \ref{fig:tinynav_route_example} illustrates one example configuration. 

\subsection{Proposed Model}

The constraints of \mbox{TFLite Micro} and \mbox{ESP-NN} limit the available neural network architectures, as both frameworks lack support for recurrent and attention-based layers such as GRU, LSTM, or multi-head attention, restricting temporal modeling of sequential depth frames. While ESP-NN provides hardware-accelerated kernels achieving up to 7$\times$ speed-ups over standard C implementations \cite{espressif2026espnn}, these are limited to specific operations, constraining design choices. At each inference step, the model processes a sliding window of 20 frames, with new frames added, and old frames discarded in shared memory between the control loop and inference cores, ensuring the network always operates on the most recent observations even under transient latency spikes.

To encode temporal information for wall avoidance and implicit speed estimation, the 20-frame sequence is represented as a single input tensor with 20 channels corresponding to time steps. Although a 3D CNN would ideally capture spatial-temporal correlations, its latency exceeds the 50~ms inference limit, so a 2D CNN operating on temporally stacked channels is used. This structure leverages the fixed frame ordering to allow the network to infer motion cues, trading some temporal expressiveness for real-time compliance.

\begin{figure}[h]
\centering
\includegraphics[width=1\columnwidth]{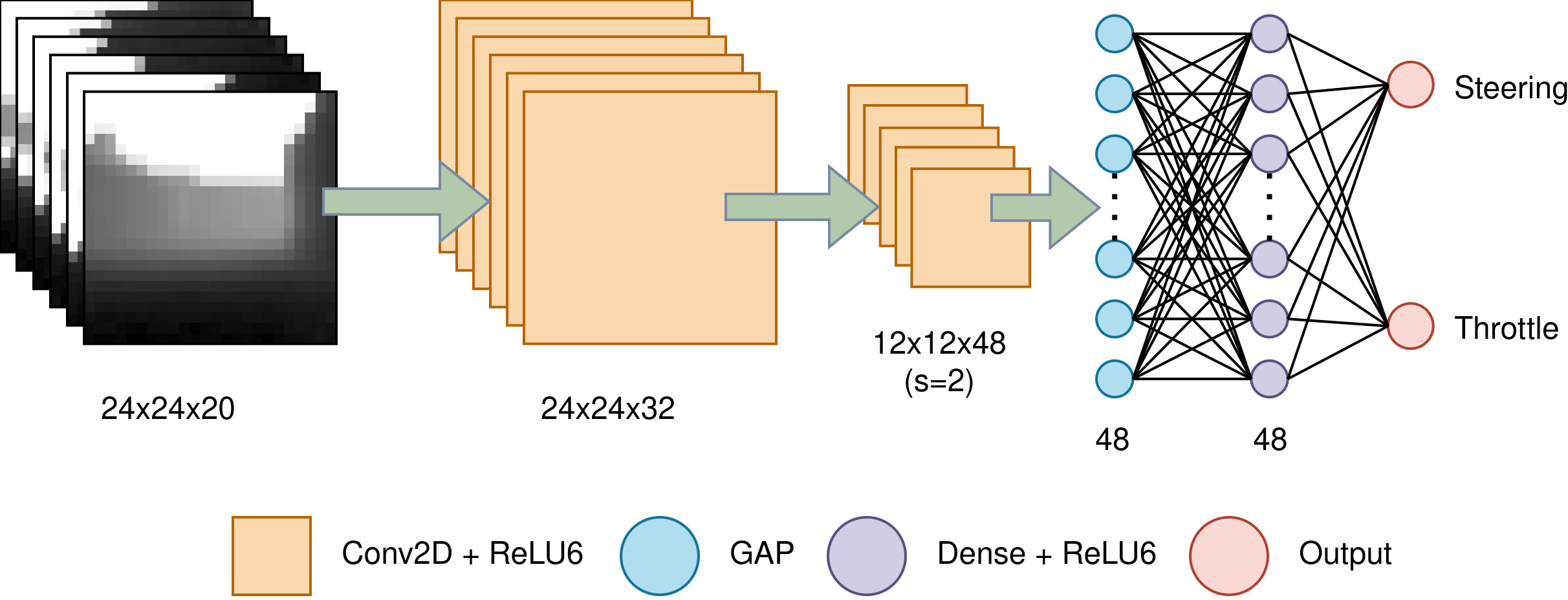}
\caption{Model Architecture}
\label{fig:tinynav_model_architecture}
\end{figure}

Dense layers follow the convolutional backbone to refine outputs without substantially increasing parameters. The network has separate output heads for steering and throttle, using appropriate activation ranges (-1 to 1 for steering, 0 to 1 for throttle) while sharing intermediate dense layers to capture the coupling between speed and steering inherent to the tank-drive platform. The resulting architecture, shown in Figure~\ref{fig:tinynav_model_architecture}, achieves 30~ms inference while only consuming 23k parameters, providing sufficient headroom to avoid latency in the control pipeline.

\begin{figure}[h]
\centering
\includegraphics[width=1\columnwidth]{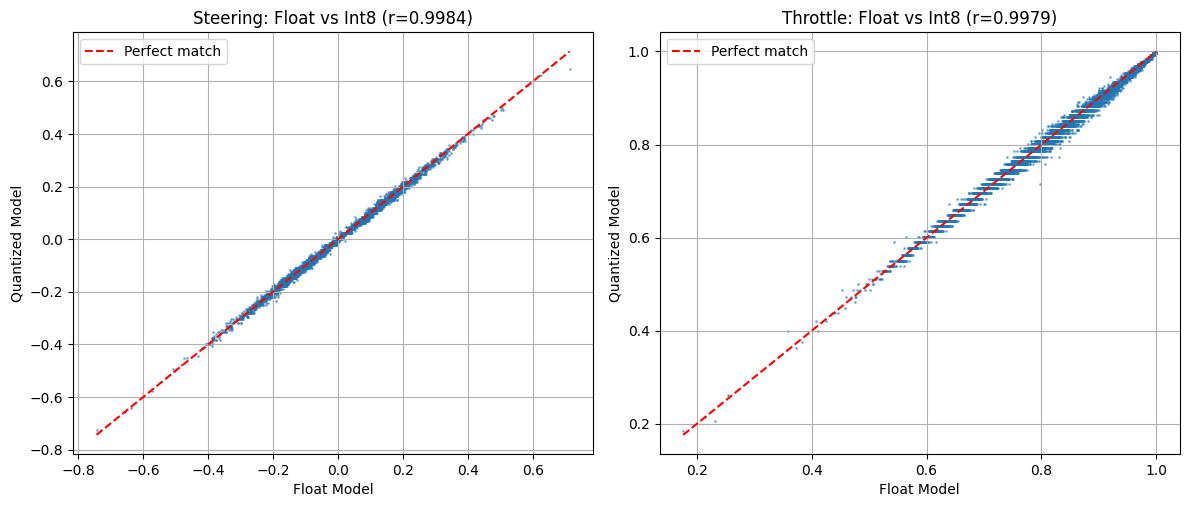}
\caption{Quantization Correlation Float vs. INT8}
\label{fig:tinynav_quantization_correlation}
\end{figure}

In order to fit the model within the memory constraints of the microcontroller and to leverage the performance advantages of integer arithmetic supported by the kernel, we applied TFLite’s built in post-training quantization with default optimization settings. A representative dataset consisting of the entire training set was used to calibrate the quantization parameters so that the activation ranges accurately reflected the true data distribution. Because the model was small, running the process over the full dataset was computationally feasible.
Figure \ref{fig:tinynav_quantization_correlation} shows the correlation between the floating point and INT8 quantized outputs for the steering and throttle heads. When evaluated on the validation set, the quantized model preserved 99.84\% of the steering accuracy and 99.79\% of the throttle accuracy, indicating that the reduction in precision resulted in negligible performance degradation while significantly improving memory efficiency and inference speed on the microcontroller.

\subsection{Validation}
\begin{figure}[h]
\centering
\includegraphics[width=1\columnwidth]{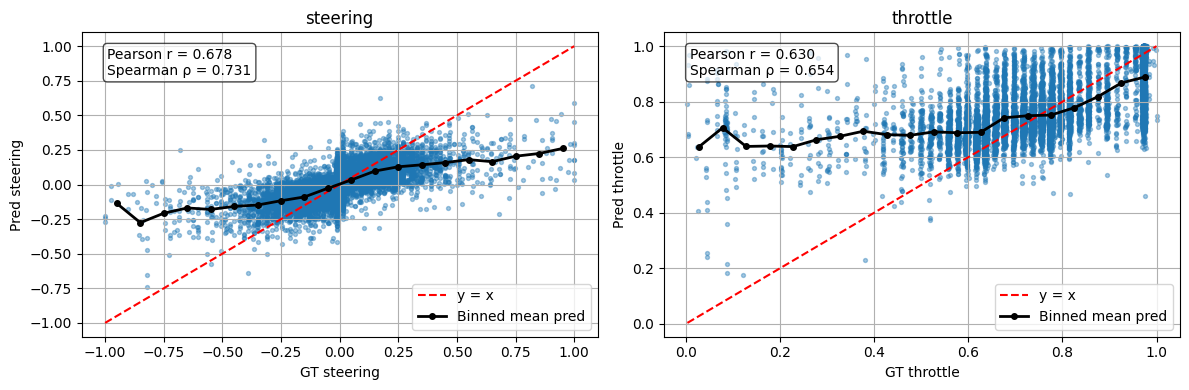}
\caption{Steering \& Throttle correlation graphs}
\label{fig:tinynav_prediction_vs_groundtruth_correlation}
\end{figure}

\begin{figure}[h]
\centering
\includegraphics[width=1\columnwidth]{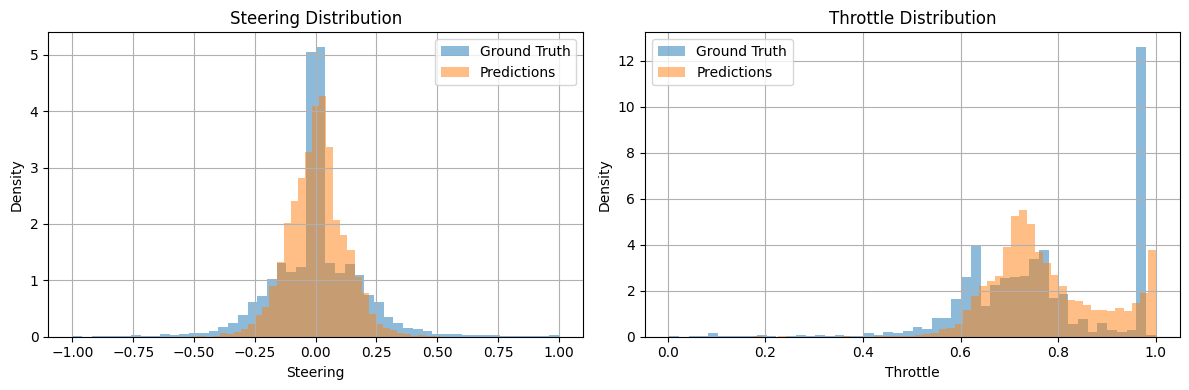}
\caption{Ground Truth vs. Prediction Distributions}
\label{fig:tinynav_prediction_vs_groundtruth_distribution}
\end{figure}

To comprehensively evaluate the model's performance and ensure reliable real-world deployment, several validation techniques were employed beyond standard scalar loss tracking:

\begin{itemize}
    \item Correlation Analysis: Figure \ref{fig:tinynav_prediction_vs_groundtruth_correlation} plots steering and throttle predicted values against ground truth (GT) targets. Binned mean predictions were overlaid to visualize the central tendency of the predictions. To quantify this relationship, both the Pearson Correlation Coefficient ($r$) and the Spearman Rank Correlation ($\rho$) were computed, measuring the linear and monotonic alignment between the network's predictions and the actual human-driven data.
    
    \item Output Distribution Matching: Figure \ref{fig:tinynav_prediction_vs_groundtruth_distribution} shows the probability density of the predictions versus the ground truth. This validation step is crucial to ensure that the model does not suffer from regression to the mean (e.g., constantly predicting a steering value of 0). The overlapping distributions verify that the model correctly predicts the full variance of necessary maneuvers, including sharp turns and full stops.
\end{itemize}

\subsection{Interpretation \& Results}
Interpreting the underlying decision-making process of CNNs remains a significant challenge. To address this, we employed Gradient-weighted Class Activation Mapping (Grad-CAM) \cite{DBLP:journals/corr/SelvarajuDVCPB16}. Grad-CAM utilizes the gradients of the steering and throttle targets flowing into the final convolutional layer to generate a coarse localization map, effectively highlighting the spatial regions most influential to the model's predictions.

\begin{figure}[h]
\centering
\includegraphics[width=1\columnwidth]{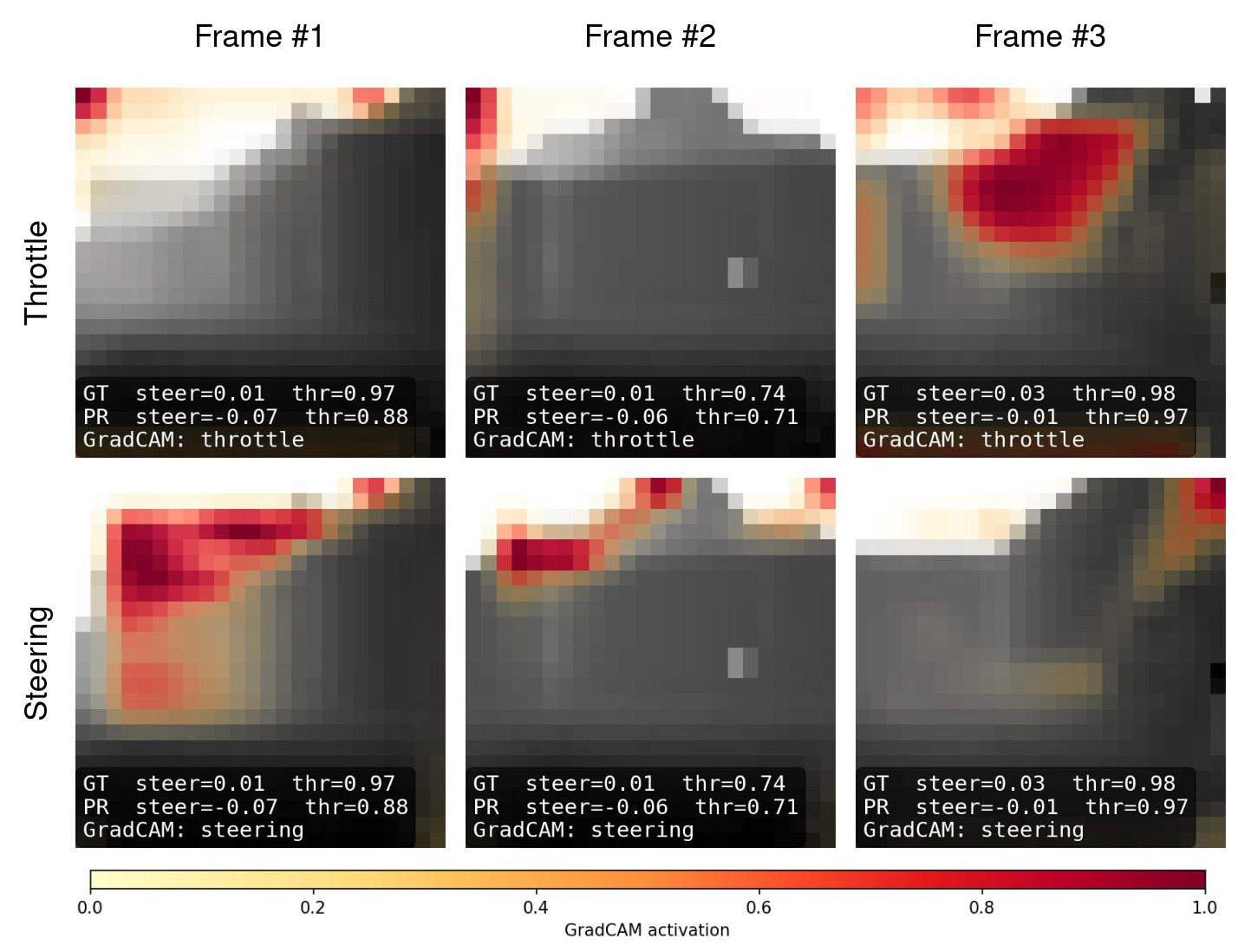}
\caption{Steering \& Throttle activation examples}
\label{fig:tinynav_steering_throttle_activations}
\end{figure}

As illustrated in Figure \ref{fig:tinynav_steering_throttle_activations}, the steering head focuses on the upper regions of the frame (the ``sky"). Because the training data excludes complex backgrounds that could misguide predictions, which falls outside the scope of this project, this upper area contains distinct, clear edges delineating the track boundaries, serving as a reliable predictive metric. Furthermore, in specific instances such as Frame \#3, the activation maps reveal that the steering head attends directly to the nearest bounding wall, indicating a positive sign of spatial awareness.

On the other hand, the throttle head exhibits localized attention mechanisms. It frequently focuses on specific contextual cues, such as the corners of the frame to identify track openings, or directly on approaching dead-ends (walls) to modulate speed appropriately. 

It is important to note that the model is not flawless. A correlation of 0.6 for the steering and throttle outputs indicates that the resulting attention regions are not perfect. Nevertheless, this analysis provides a valuable understanding of what features the current training dataset prioritizes and successfully highlights critical areas for improvement in future data collection efforts.

Empirical evaluation was conducted on both training-like layouts and unseen track configurations. On simple track geometries similar to those used during training, the robot completed 40 consecutive laps without collision. Across these runs, steering and throttle outputs remained stable, demonstrating reliable closed-loop behavior under familiar structural conditions.

On more complex layouts outside the training and validation sets, full circuits were achieved, though not consistently and with occasional minor wall contacts. Despite this, the robot maintained forward motion without external intervention.

Throttle predictions vary according to track structure. On long straight segments, throttle values approached their upper range, while near dead ends or sharp turns, throttle decreased smoothly, in some cases nearing zero before the maneuver. This behavior emerges directly from the learned depth-to-control mapping.

\section{Limitations}

Despite the numerous advantages and opportunities provided by AI applications on microcontroller platforms, its limiting hardware still presents notable technical constraints to AI applications. To begin, microcontroller technology holds limits to ideal parameter counts for machine learning models. Parameter count refers to the quantity of learned variables within a neural network, including weights and biases across all layers. ESP32 microcontrollers can run up to 50,000 parameters at 20 frames per second depending on the architecture, and up to 500,000 parameters if inference time is not a significant issue. This presents scalability concerns with model complexity, preventing deployment of deeper and more computationally intensive architectures.

An additional challenge to microcontroller-AI applications is memory. Unlike standard laptop or desktop devices, which oftentimes contain several GBs of RAM, microcontrollers possess minute amounts of memory.  Even the higher-quality ESP-32 microcontroller devices come up to 32MB of RAM. Subsequently, this heavily restricts the size of deployable models, as it diminishes a system's pattern recognition and predictive modelling capacities. Consequently, AI systems must be rigorously optimized on microcontroller software at the potential expense of peak performance.
    
Moreover, microcontrollers often exhibit limited connectivity capabilities when peripheral interfaces contend for shared hardware resources. For example, certain configurations prevent the simultaneous use of an SD card module and a Wi-Fi module because both rely on the same communication lines (e.g., SPI buses) to support high data throughput. This resource conflict constrains concurrent operation and reduces overall system flexibility, particularly in applications requiring both local storage and network connectivity. In a world that's increasingly becoming more and more dependent on devices with real-time cloud networks and constant communication, this significantly narrows application possibilities with concerns for flexibility and interconnectivity. 

Another key limitation is the lack of diversity in the collected datasets for the model training process. Although the robot was driven along different paths by different people, these paths shared nearly identical structural properties. Firstly, the model is not openly exposed to situations requiring frequent complex or rapid steering adjustments, so drifting might still be a possibility. Furthermore, the wall geometry, path width, and obstacle height remained constant to simplify the scope of the project. These constraints arose from the fixed orientation and limited vertical sensing range of the depth camera, which prevented the robot from detecting objects at significantly different heights or scales. As a result, the robot was only exposed to a narrow set of obstacle types, primarily consisting of walls of shared height and occasional backpacks placed along the route. The resulting model is not capable of generalizing to new environments due to this restricted exposure.

TinyNav’s autonomous driving is currently limited to forward motion and left-right steering. In contrast, driving backward introduces far more challenging dynamics. Reverse motion is inherently less stable, with highly sensitive and nonlinear behavior that requires a significantly richer and more diverse training dataset. As a result, attempting to turn while driving backward greatly increases the likelihood that the vehicle will diverge from its intended path. 

TinyNav is limited to using the depth camera as its only source of motion‑related information. The robot does not incorporate wheel encoders to measure wheel rotations, speed, or distance travelled, leaving it without true odometry or low‑level movement feedback. As a result, the model plans trajectories based solely on detected or predicted obstacles at a fixed height, rather than on how the robot is actually moving through space. Relying exclusively on feature tracking and depth variation introduces additional noise and drift, especially under low‑texture environments, making state estimation less reliable. Incorporating encoder feedback would provide the closed‑loop motion data needed for the CNN to generate more stable, consistent, and predictable control outputs.

\section{Conclusion}

TinyNav demonstrates that intelligent AI systems can be successfully deployed on constrained hardware, without reliance on high-performance computational resources. Through thorough data collection, data refining, rigorous model training and optimization, TinyNav displays that a machine learning algorithm can be implemented on hardware with limitations in memory, processing power, and parameter count.

With AI becoming extremely dominant in today's world, it is imperative that we begin to pay more attention to more efficient and cost-effective hardware for AI models. TinyNav highlights an approach to devices with reduced latency, lower energy consumption, and more independent systems. This provides opportunities for improved and re-imagined drones, cars, and other mobile robotic systems. While issues such as scalability, memory, and connectivity handicap TinyML, further developments in model compression techniques and expanding datasets can enhance performance with the given hardware constraints. Despite this, TinyNav proves that not only is it tangible to deploy such a model on limited hardware, it's a clear next step for more accessible, energy-efficient AI systems and robotics.

\section{Reproducibility}

All code, models, datasets, and hardware documentation required to reproduce this project are publicly available:

\begin{itemize}
    \item CNN training code (model architecture, preprocessing, and training scripts).\footnote{\url{https://github.com/regularpooria/tinynav_cnn}}
    
    \item Robot firmware implemented using ESP-IDF.\footnote{\url{https://github.com/regularpooria/TinyNav}}
    
    \item Collected depth camera dataset used for training and evaluation.\footnote{\url{https://huggingface.co/datasets/regularpooria/tinynav_depth_camera_circuits}}
    
    \item Complete hardware bill of materials (BOM) and component order list.\footnote{\url{https://github.com/regularpooria/TinyNav/blob/main/images_videos/TinyNav_order_list.pdf}}
\end{itemize}

\section{Acknowledgements}
Queen's University's AI Club (QMIND\footnote{\url{https://qmind.ca}}) provided the financial support necessary to procure the components required for the development and implementation of this project.

\newpage
\bibliographystyle{IEEEtran}
\bibliography{references}

\end{document}